\documentclass[10.5pt,compsoc]{BD}
\usepackage{graphicx}
\usepackage{footmisc}
\usepackage{subfigure}
\usepackage{url}
\usepackage{multirow}
\usepackage[noadjust]{cite}
\usepackage{amsmath,amsthm}
\usepackage{amssymb,amsfonts}
\usepackage{booktabs}
\usepackage{color}
\usepackage{ccaption}
\usepackage{booktabs}
\usepackage{float}
\usepackage{fancyhdr}
\usepackage{caption}
\usepackage{xcolor,stfloats}
\usepackage{comment}
\graphicspath{{figures/}}
\usepackage{cuted}  
\usepackage{captionhack}
\usepackage{epstopdf}

\headevenname{\zihao{-5}{\textbf{\emph{Big Data Mining and Analytics, January}}} 2018, 1(1): 000-000}%
\headoddname{{\sf First author et al.:}\quad {\textbf{\emph{Click and Type Your Title Here ...}}}}%

\newtheoremstyle{mystyle}{0pt}{0pt}{\normalfont}{1em}{\bf}{}{1em}{}
\theoremstyle{mystyle}
\renewcommand\figurename{Fig.~}

\newcommand{\nop}[1]{}

\makeatletter
\renewcommand{\@biblabel}[1]{[#1]\hfill}
\makeatother

\begin{document}

\thispagestyle{empty}



\hyphenpenalty=50000

\makeatletter
\newcommand\mysmall{\@setfontsize\mysmall{7}{9.5}}

\newenvironment{tablehere}
  {\def\@captype{table}}
  {}
\newenvironment{figurehere}
  {\def\@captype{figure}}
  {}

\thispagestyle{plain}%
\thispagestyle{empty}%

\let\temp\footnote
\renewcommand \footnote[1]{\temp{\zihao{-5}#1}}
{}
\vspace*{-40pt}
\noindent{\zihao{5-}\textbf{\scalebox{0.95}[1.0]{\makebox[5.9cm][s]
{BIG\hfill  DATA \hfill MINING \hfill AND \hfill ANALYTICS}}}}

\vskip .2mm
{\zihao{5-}
\textbf{
\hspace{-5mm}
\scalebox{1}[1.0]{\makebox[5.6cm][s]{%
I\hspace{0.70pt}S\hspace{0.70pt}S\hspace{0.70pt}N\hspace{0.70pt}{\color{white}%
2\hspace{-2pt}2\hspace{0.70pt}}2\hspace{0.70pt}0\hspace{0.70pt}9\hspace{0.70pt}6\hspace{0.70pt}-\hspace{0.70pt%
}0\hspace{0.70pt}6\hspace{0.70pt}5\hspace{0.00pt}4\hspace{0.70pt}\hspace{0.70pt%
}\hspace{0.70pt}{\color{white}l\hspace{0.70pt}l\hspace{0.70pt}}0\hspace{0.70pt}%
?\hspace{0.70pt}/\hspace{0.70pt}?\hspace{0.70pt}?\hspace{0.70pt}{\color{white}%
l\hspace{0.70pt}l\hspace{0.70pt}}p\hspace{0.70pt}p\hspace{0.70pt}?\hspace{0.70pt}?\hspace{0.70pt}?%
--\hspace{ 0.70pt}?\hspace{0.70pt}?\hspace{0.70pt}?}}}}

\vskip .2mm\noindent
{\zihao{5-}\textbf{\scalebox{1}[1.0]{\makebox[5.6cm][s]{%
V\hspace{0.4pt}o\hspace{0.4pt}l\hspace{0.4pt}u\hspace{0.4pt}m\hspace{0.4pt}%
e\hspace{0.4em}1\hspace{0.4pt},\hspace{0.8em}N\hspace{0.4pt}u\hspace{0.4pt}%
m\hspace{0.4pt}b\hspace{0.4pt}e\hspace{0.4pt}r\hspace{0.4em}1,\hspace{0.8em}%
J\hspace{0.4pt}a\hspace{0.4pt}n\hspace{0.4pt}u\hspace{0.4pt}a\hspace{0.4pt}%
\hspace{0.4pt}r\hspace{0.4pt}y\hspace{0.4em}2\hspace{0.4pt}0\hspace{0.4pt}1\hspace{0.4pt}8}}}}

\vskip .2mm\noindent
{\zihao{5-}\textbf{\scalebox{1}[1.0]{\makebox[5.6cm][s]{%
\color{white}{V\hfill o\hfill l\hfill u\hfill m\hfill%
e\hspace{0.356em}1,\hspace{0.356em}N\hfill u\hfill%
m\hfill b\hfill e\hfill r\hspace{0.356em}1,\hspace{0.356em}%
S\hfill e\hfill p\hfill t\hfill e\hfill%
m\hfill b\hfill e\hfil lr\hspace{0.356em}2\hfill0\hfill1\hfill8}}}}}\\

\begin{strip}
{\center
{\zihao{3}\textbf{
Recent Advances in Federated Learning Driven Large Language Models: A Survey on Architecture, Performance, and Security}}
\vskip 9mm}

{\center {\sf \zihao{5}
Youyang Qu, Ming Liu, Tianqing Zhu, Longxiang Gao, Shui Yu, and Wanlei Zhou
}
\vskip 5mm}
%

\centering{
\begin{tabular}{p{160mm}}

{ \zihao{-5}
\linespread{1.6667} %
\noindent
\bf{Abstract:} {\sf
Federated Learning (FL) offers a promising paradigm for training Large Language Models (LLMs) in a decentralized manner while preserving data privacy and minimizing communication overhead. This survey examines recent advancements in FL-driven LLMs, with a particular emphasis on architectural designs, performance optimization, and security concerns, including the emerging area of machine unlearning. In this context, machine unlearning refers to the systematic removal of specific data contributions from trained models to comply with privacy regulations such as the Right to be Forgotten. We review a range of strategies enabling unlearning in federated LLMs, including perturbation-based methods, model decomposition, and incremental retraining, while evaluating their trade-offs in terms of efficiency, privacy guarantees, and model utility. Through selected case studies and empirical evaluations, we analyze how these methods perform in practical FL scenarios. This survey identifies critical research directions toward developing secure, adaptable, and high-performing federated LLM systems for real-world deployment.}
}

\end{tabular}
}
\vskip 6mm

\vskip -3mm
\zihao{6}

\end{strip}

\thispagestyle{plain}
\thispagestyle{empty}%
\makeatother
\pagestyle{tstheadings}

\vspace{3.5mm}


\section{Introduction}
\label{s:introduction}
\noindent
The rapid evolution of artificial intelligence has led to significant breakthroughs across various domains, particularly with the advent of Large Language Models (LLMs) such as GPT~\cite{zheng2025review,radford2019language} and BERT~\cite{devlin2018bert,gardazi2025bert}. These models have revolutionized machine understanding and generation of human language, enabling applications ranging from automated customer support to advanced natural language processing tasks~\cite{shanahan2023role,sun2024scieval}. As LLMs continue to scale, the demand for decentralized, privacy-preserving training frameworks has grown, especially in sensitive domains like chemistry~\cite{zheng2025large}, healthcare~\cite{gallifant2025tripod}, education~\cite{raihan2025large}, and finance~\cite{jha2025does}, where data sharing is often constrained by legal and ethical considerations~\cite{lund2023chatgpt}.

Federated Learning (FL) has emerged as a promising paradigm that enables collaborative training of LLMs across multiple decentralized clients without transferring raw data~\cite{ye2024fedllm,he2023three,zhang2023fedpetuning}. This approach addresses key challenges related to data privacy~\cite{yazdinejad2024robust}, communication efficiency~\cite{deng2024communication}, and compliance with data protection regulations. FL-driven LLMs offer the potential to leverage diverse data sources while preserving individual user privacy, enhancing generalization, and ensuring robustness to data heterogeneity~\cite{chai2024survey,wang2024aggregation,huang2024federated}.

Despite its promise, the integration of FL with LLMs raises several research challenges. These include handling non-IID (non-independent and identically distributed) data~\cite{wu2024fedbiot}, ensuring convergence and consistency across clients~\cite{ling2024convergence}, minimizing communication overhead, and safeguarding models against privacy attacks such as membership inference and gradient leakage~\cite{hu2024federated}. Moreover, the growing emphasis on data sovereignty has brought forward the need for mechanisms like machine unlearning~\cite{qu2025frontier}, which allows the removal of individual data contributions from models to comply with policies such as the Right to be Forgotten~\cite{gdpr_2016}.

Motivated by these emerging needs, this survey aims to provide a comprehensive and structured overview of the current research landscape surrounding Federated Learning for Large Language Models, with an emphasis on architecture, performance optimization, and security and privacy considerations. We highlight recent innovations, practical implementations, and open research challenges at the intersection of federated optimization and large-scale language modeling. The main contributions of this survey are as follows.

\begin{itemize} 
    \item \textbf{Comprehensive Overview of Architectures:} We present a taxonomy of FL-driven LLM system architectures, outlining design patterns, coordination schemes (centralized, decentralized, hybrid), and training workflows tailored for privacy-preserving large-scale NLP tasks.
    \item \textbf{Analysis of Performance and Adaptability:} We assess how FL impacts the training dynamics, convergence speed, and generalization ability of LLMs across heterogeneous environments, and explore optimization techniques such as model compression, personalization, and communication-efficient training.
    \item \textbf{Security and Privacy Perspectives:} We review emerging threats to FL-LLMs, such as privacy leakage and poisoning attacks, and survey mitigation techniques including secure aggregation, differential privacy, and machine unlearning strategies.
\end{itemize}

The remainder of the survey is as follows. Section 2 provides a detailed exploration of the foundations of FL and its integration with LLM, highlighting the synergies and potential enhancements. Section 3 examines the specific adaptations and innovations in FL algorithms that facilitate efficient and effective training of LLMs across distributed agents. Section 4 presents a critical analysis of current implementations and discusses the technical challenges, practical limitations, and potential solutions. Section 5 concludes the survey with a discussion on future research directions and the broader implications of federated LLMs in various emerging domains.

\begin{figure}
\label{fig:workflow}
  \centering
  \includegraphics[width=0.40\textwidth]{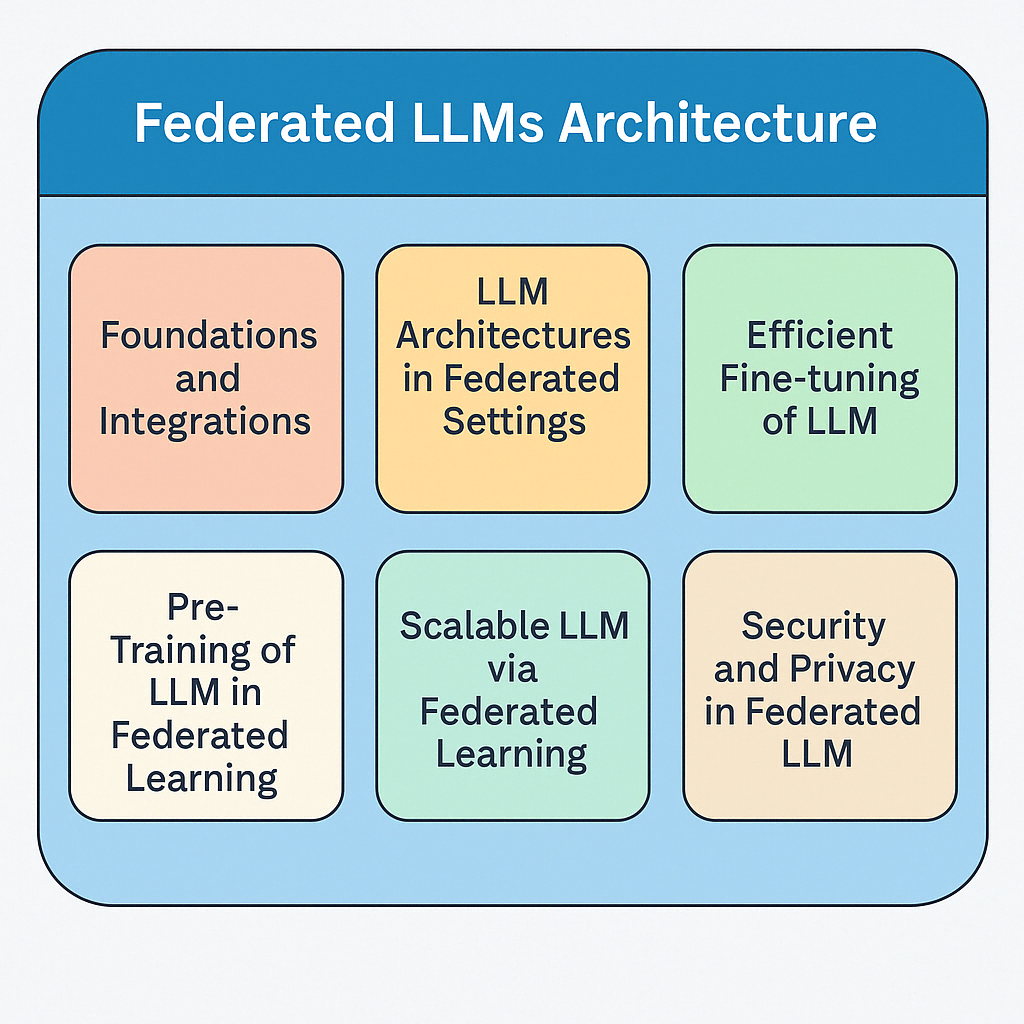}
  \caption{Organization of this survey}
\end{figure}

\section{Foundations of Federated Learning and Large Language Models}
\label{section:foundations}

This section provides the foundational background necessary to understand the interplay between Federated Learning (FL) and Large Language Models (LLMs). We first explore the basic principles and motivations of FL, then outline the evolution and architecture of LLMs, and finally describe how the two can be effectively integrated for privacy-preserving and distributed AI solutions.

\subsection{Federated Learning: Principles \& Motivations}

Federated Learning is a decentralized learning paradigm that enables multiple clients, such as edge devices or organizations, to collaboratively train a shared model without exposing their raw data~\cite{beltran2023decentralized,liu2024vertical}. This is particularly important in domains where privacy, data ownership, or regulatory constraints prevent centralizing datasets~\cite{wen2023survey}. Unlike traditional machine learning pipelines that aggregate data into a central server, FL operates by distributing the learning process~\cite{ye2023heterogeneous}. Each client trains a local model on its private data and then shares model updates, such as gradients or weights, with a coordinating server that aggregates them to update the global model~\cite{tan2022towards}.

This approach offers several benefits. First, it enhances privacy by keeping data local~\cite{huang2023rethinking}. Second, it supports scalability across heterogeneous and distributed environments~\cite{huang2024federated}. Third, it provides a viable pathway for compliance with data protection regulations such as GDPR or CCPA~\cite{ccpa_2018}. Finally, FL enables local personalization, where clients can adapt the shared model to suit specific needs or user preferences~\cite{pillutla2022federated}. These features make FL a compelling solution for privacy-sensitive applications in healthcare, finance, mobile computing, and other sectors.

\subsection{Fundamentals of Large Language Models}

As FL provides a secure and distributed framework for training, the rise of Large Language Models introduces a new opportunity to apply it to models with unprecedented capability. LLMs, such as GPT~\cite{zhang2025rank,radford2019language} and BERT~\cite{hou2025tf,devlin2018bert}, are deep neural architectures trained on vast textual corpora using unsupervised or self-supervised learning objectives. Most modern LLMs rely on the Transformer architecture~\cite{feng2025dit4edit,han2021transformer}, which introduced the self-attention mechanism that allows models to capture dependencies across arbitrary-length sequences. This design drastically improves upon prior models like RNNs and LSTMs in terms of scalability and contextual understanding~\cite{shiri2023comprehensive}.

Pretraining tasks such as next-token prediction~\cite{li2024mechanics} or masked language modeling~\cite{jin2024wordtransabsa} enable these models to develop a strong general-purpose understanding of language, which can be fine-tuned with relatively small task-specific datasets. Over the years, LLMs have grown from millions to hundreds of billions of parameters~\cite{chang2024survey}, expanding their ability to perform a wide range of language tasks—from summarization and translation to reasoning and code generation~\cite{xie2024doremi,min2023recent}. However, this growth comes with challenges: training and deploying LLMs is computationally expensive and typically requires significant infrastructure and engineering effort~\cite{xu2024unleashing}.

\subsection{Integration of Federated Learning and LLMs}

The convergence of FL and LLMs creates a powerful framework that supports training and deploying large-scale models across data silos without compromising privacy~\cite{ye2024fedllm}. The motivation for integrating these two technologies stems from the need to utilize sensitive or siloed datasets, such as clinical notes in hospitals~\cite{peng2024depth} or user interactions on mobile devices~\cite{qu2025mobile}, while adhering to strict privacy constraints. FL provides the decentralized mechanism for training, while LLMs offer a general-purpose, high-capacity architecture capable of adapting to diverse linguistic tasks~\cite{xu2025flexfl}.

Implementing FL for LLMs introduces a number of technical challenges. These include handling heterogeneous data distributions across clients, reducing the communication overhead of model updates, mitigating privacy risks associated with gradient leakage, and achieving consistent convergence despite partial participation~\cite{ye2024openfedllm,mawela2025web}. Additionally, there is growing interest in mechanisms like machine unlearning within FL settings, allowing models to forget specific data contributions post-training, which is crucial for compliance with emerging privacy laws~\cite{rashid2025forget}.

Real-world applications of FL-LLMs are already being explored in sectors such as healthcare, finance, and smart devices. Hospitals can jointly train medical LLMs without sharing patient data, financial institutions can build collaborative assistants while retaining proprietary transaction logs, and edge devices can personalize LLMs locally for user-specific tasks like voice input or text prediction~\cite{yang2023cancersqa, kheddar2024automatic}. These scenarios highlight the relevance and promise of FL-LLMs for building secure, personalized, and scalable language-based systems.

\begin{figure}[ht]
  \centering
  \includegraphics[width=0.48\textwidth]{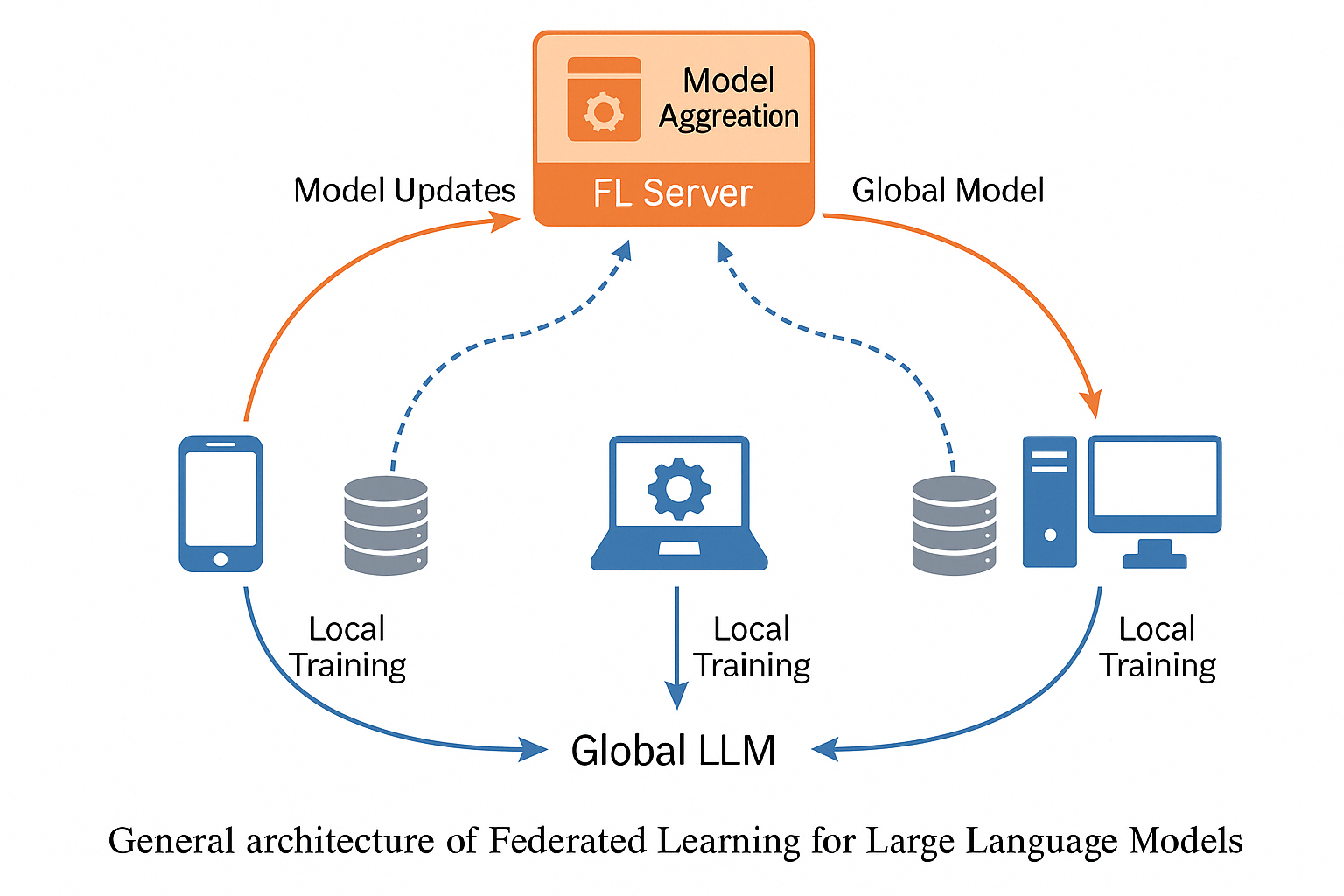}
  \caption{General architecture of Federated Learning for Large Language Models}
  \label{fig:architecutre}
\end{figure}

\section{Federated Large Language Models}
\label{section:FL-LLM}

In this section, we present up-to-date research on federated LLMs from the following perspectives: LLM architectures in federated settings, efficient fine-tuning of federated LLM, pre-training of LLM in federated learning, scalable LLM via federated learning, and security \& privacy in federated LLM.

\subsection{Federated LLM architectures}

\begin{table*}
\centering
\caption{Overview of Federated Large Language Model Architectures}
\vskip 6mm
\label{tab:my_label}
\begin{tabular}{c|p{4cm}|p{2cm}|p{2cm}|p{2cm}|p{4cm}}
\toprule
\textbf{No.} & \textbf{Contributions} & \textbf{Technique} & \textbf{LM used} & \textbf{Dataset} & \textbf{Limitations} \\ 
\midrule
1 & Implementation of N-gram models in federated settings, dealing with distributed data. & Federated learning & N-gram RNN & Not specified & High communication overhead, data privacy concerns. \\ 
\midrule
2 & Enhances privacy and reduces server communication through localized computations. & Federated Reconstruction & LSTM-based & MovieLens  & Complexity in local computations, potential performance trade-offs. \\ 
\midrule
3 & Framework for training and deploying LLMs in federated settings, addressing scalability and efficiency. & FederatedScope & LLaMa-7B & HumanEval, HELM, etc & Scalability issues, computational demands. \\ 
\midrule
4 & Integration of LLMs with edge computing for autonomous decision-making in connected environments. & Edge AI, Federated Learning & GPT-3 & Synthetic challenging user request dataset & Resource constraints on edge devices, deployment challenges. \\ 
\midrule
5 & Techniques for reducing LLM parameter count in federated settings, ensuring efficiency. & Model pruning, Knowledge distillation & Bert-based & IMDP, Yelp & Compromise on model performance, complexity in model reduction techniques. \\ 
\bottomrule
\end{tabular}
\end{table*}

Chen et al.~\cite{chen2019federated} explore the adaptation of N-gram language models to federated learning environments. It discusses the challenges of maintaining model performance when training data is distributed and not centrally stored. The primary focus is on how to effectively aggregate learned N-grams from multiple nodes while preserving privacy and minimizing communication overhead.

Singhal et al.~\cite{singhal2021federated} introduce a novel approach named ``Federated Reconstruction,'' which enhances privacy in federated learning by allowing more computation to be performed locally. The paper particularly emphasizes its application to large language models, discussing how it can reduce the amount of data required to be shared with the server, thus enhancing data privacy and reducing bandwidth requirements.

Kuang et al.~\cite{kuang2023federatedscope} present FEDERATEDSCOPE, a framework designed to facilitate the training and deployment of large language models in a federated setting. It offers insights into the architectural adjustments and optimizations necessary to handle the complexities associated with training large models across decentralized data sources. The paper also explores scalability and efficiency challenges, providing solutions to maintain robust model performance.

Shen et al.~\cite{shen2024large} discusses the integration of large language models (LLMs) with edge computing devices in a federated learning context. It focuses on how LLMs can empower edge devices for autonomous decision-making, highlighting the potential for LLMs to enhance connected intelligence across various applications. The paper addresses the challenges of deploying complex models on resource-constrained edge devices.

Focusing on the issue of model size, Jiang et al.~\cite{jiang2023low} propose techniques for reducing the parameter count of large language models in federated settings. It explores methods like model pruning and knowledge distillation to maintain model efficacy with fewer parameters, which is critical for efficient data transmission and quick adaptation in federated networks.

\begin{figure*}
\label{fig:finetune}
  \centering
  \includegraphics[width=0.80\textwidth]{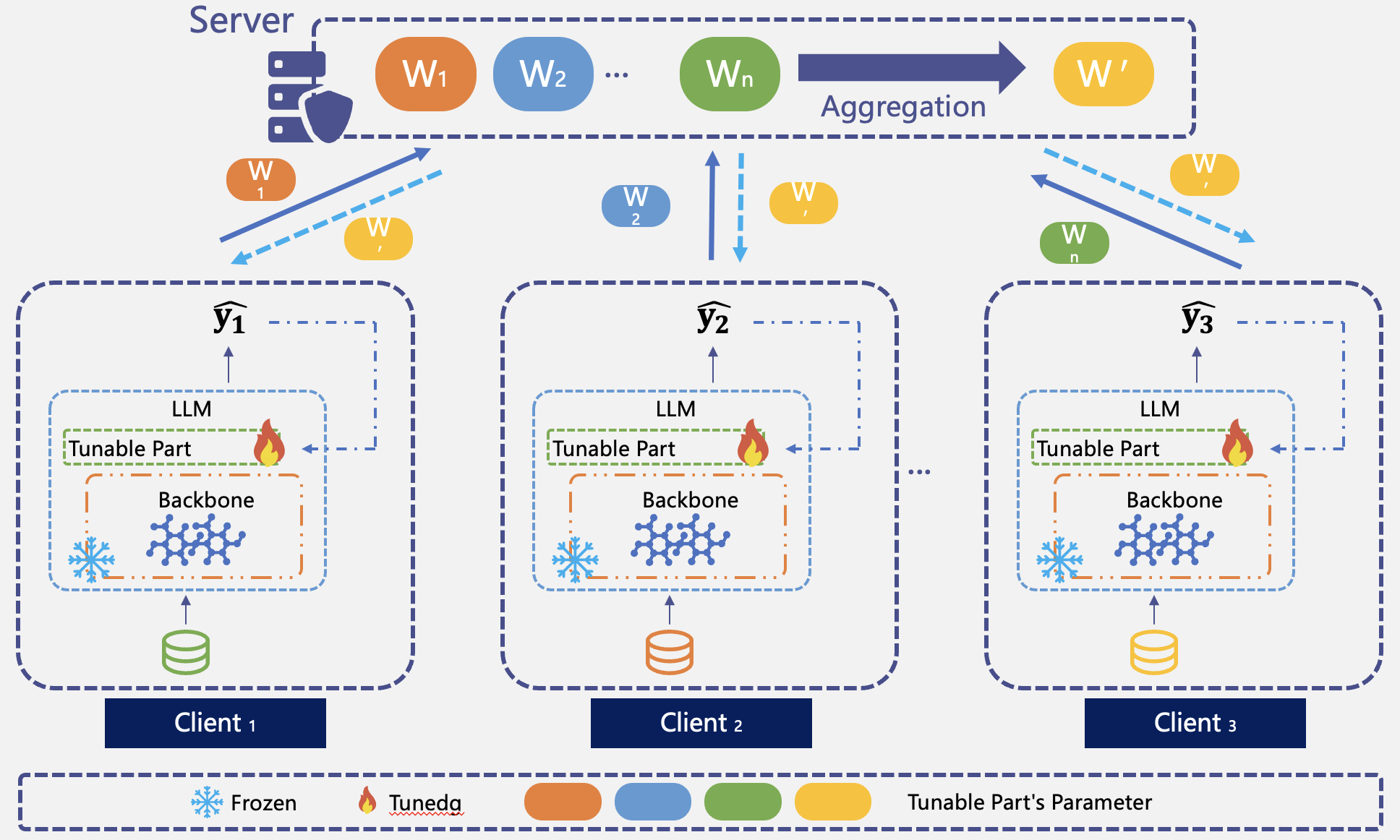}
  \caption{Fine-tuning of Federated LLMs for Swarm Intelligence }
\end{figure*}

These papers collectively cover a spectrum of considerations for deploying large language models in federated learning environments. Starting from the basic implementation of N-gram models in a federated manner, the survey transitions to advanced strategies like Federated Reconstruction for enhancing privacy and computational efficiency. It then moves into a discussion on frameworks and architectural adjustments necessary for scaling LLMs across decentralized networks. The integration with edge AI represents an application-focused advancement, providing a practical viewpoint on deploying reduced-parameter models in real-world settings. This logical progression from foundational concepts to practical implementations and optimizations illustrates a comprehensive view of current research in the domain of federated large language models.

The integration of large language models (LLMs) in federated settings, as explored in these papers, significantly advances the field of swarm intelligence by leveraging the collective learning capabilities of decentralized networks. The first paper's exploration of N-gram models in federated environments lays the foundational understanding of how individual nodes, like agents in a swarm, can contribute localized knowledge to build a cohesive and comprehensive model. The introduction of~\cite{chen2019federated} in the second paper enhances this concept by minimizing central coordination, thereby promoting more autonomous local decision-making to the independent yet coordinated behavior seen in natural swarms. The ~\cite{singhal2021federated} framework further capitalizes on this by scaling up the capabilities of LLMs across diverse and distributed datasets, mirroring how swarms adapt to varied environments. \cite{kuang2023federatedscope} brings these concepts into the realm of edge computing, where LLMs empower edge devices to function like intelligent agents that perform tasks and make decisions in real time, enhancing the swarm’s overall responsiveness and flexibility. Lastly, the techniques discussed in~\cite{shen2024large} for reducing the parameter count of LLMs ensure that the computational load is manageable even in resource-constrained environments, which is crucial for maintaining the efficiency and agility of a swarm. Together, these studies demonstrate how federated learning models can emulate and enhance swarm-like intelligence, promoting robust, scalable, and decentralized problem-solving capabilities in artificial systems.

\subsection{Efficient Fine-tuning of LLM}
\begin{table*}[htbp]
\centering
\caption{Summary of Efficient Fine-Tuning of LLMs in Federated Settings}
\vskip 6mm
\label{tab:efficient}
\begin{tabular}{c|p{4cm}|p{2cm}|p{2cm}|p{2cm}|p{4cm}}
\toprule
\textbf{No.} & \textbf{Contributions} & \textbf{Technique} & \textbf{LM used} & \textbf{Dataset} & \textbf{Limitations} \\ 
\midrule
1 & Integrates differential privacy efficiently in federated LLM training with minimal performance loss. & Differential privacy, noise addition, parameter clipping & CIFG 19M, Transformer 21M, etc. & WordPiece & Potential impact on model accuracy, complexity in tuning privacy parameters. \\ 
\midrule
2 & Utilizes prompt tuning and adaptive optimization to reduce parameters and adapt learning rates dynamically. & Prompt tuning, adaptive optimization & GPT-2, LLaMa, etc. & MRPC & May not reach the full model's potential, depends heavily on prompt design. \\ 
\midrule
3 & Optimizes prompt tuning in federated settings with synchronization for stable and accurate model updates. & Federated prompt tuning, synchronization mechanisms & Roberta & GLUE & Challenges in synchronization across diverse networks, prompt design limitations. \\ 
\midrule
4 & Benchmarks federated learning models and systems at scale, identifying and addressing bottlenecks. & Benchmarking, model compression, robust aggregation methods & Albert & Reddit & Scaling issues, may require significant computational resources. \\ 
\midrule
5 & Reduces data transmission with forward gradient method, enhancing privacy and efficiency. & Forward gradient technique, efficient gradient aggregation & LLaMa, Roberta, etc. & AGNEWS, Yelp, etc. & Potential loss of gradient information, requires careful implementation. \\
\bottomrule
\end{tabular}
\end{table*}

Ro et al.~\cite{ro2024efficient} address the challenges of applying differential privacy in federated settings, particularly when fine-tuning large language models (LLMs). It introduces a novel architecture that balances the trade-offs between privacy, efficiency, and performance. The proposed model uses lightweight mechanisms to ensure differential privacy without significantly degrading the model's utility. Key techniques include noise addition and parameter clipping during the training process to maintain privacy. The architecture is designed to be modular, allowing easy integration with existing federated learning frameworks. This study is pivotal as it demonstrates how privacy considerations can be seamlessly integrated into the federated fine-tuning of LLMs, ensuring that user data remains confidential while still contributing to the collective learning process.

Che et al.~\cite{che2023federated} explore the use of prompt tuning as a parameter-efficient method for fine-tuning large language models in federated learning environments. Prompt tuning involves adjusting a small set of parameters (prompts) while keeping the majority of the model fixed, significantly reducing the computational and communication overhead typically associated with training large models. Additionally, the paper introduces an adaptive optimization technique that dynamically adjusts learning rates based on the network conditions and convergence rates of the participating nodes. This approach not only enhances the efficiency of the federated learning process but also ensures that the model adapts optimally across diverse and potentially noisy datasets.

\begin{figure*}
\label{fig:pretrain}
  \centering
  \includegraphics[width=0.80\textwidth]{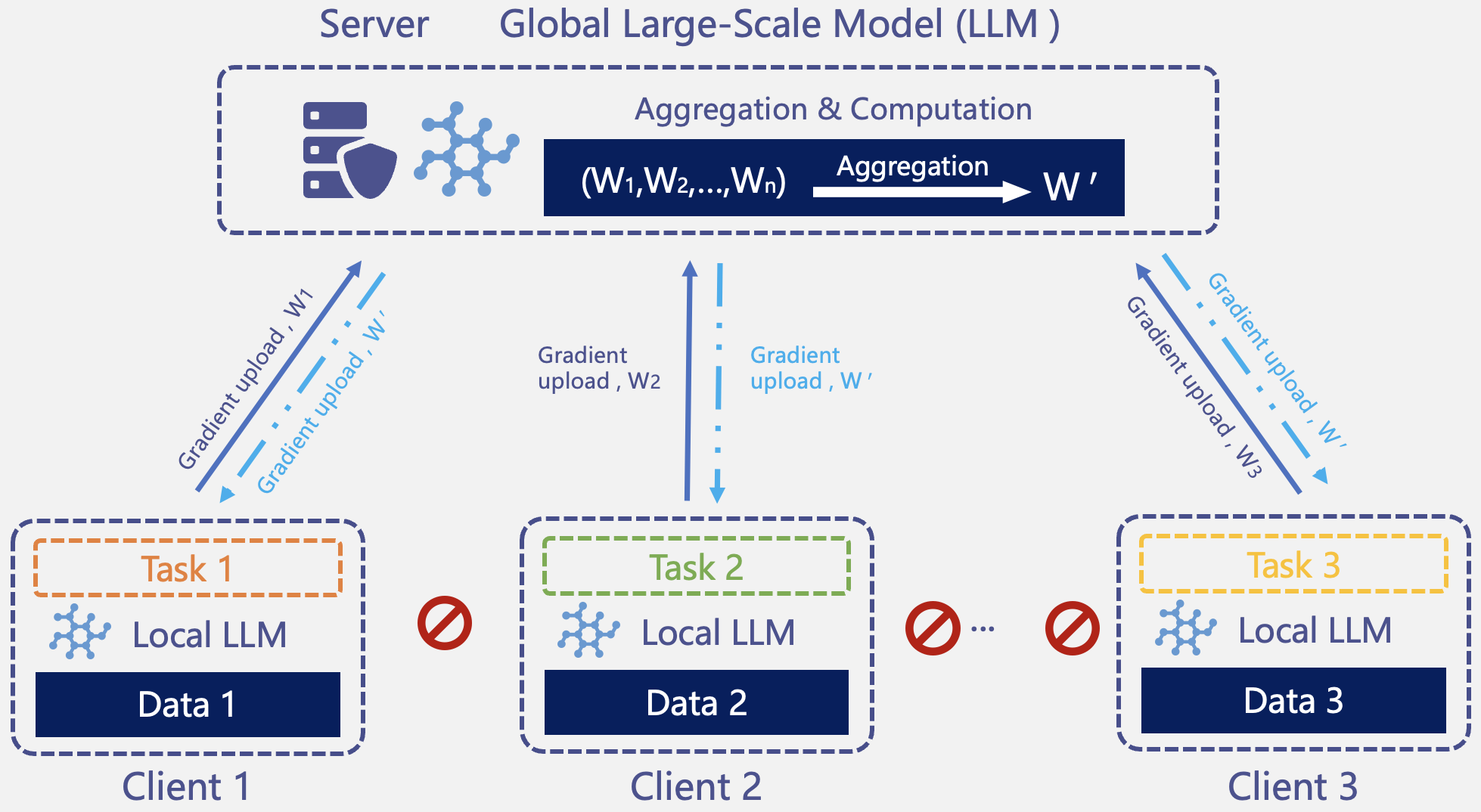}
  \caption{Pre-training of Federated LLMs for Swarm Intelligence }
\end{figure*}

Building on the concept of prompt tuning, Zhang et al.~\cite{zhang2023fedpetuning} specifically tailor this technique for federated settings. The paper presents a framework that optimizes the deployment of prompt tuning across a distributed network, ensuring that all nodes contribute effectively to the model's learning process without overwhelming the network's bandwidth. The framework also includes a novel synchronization mechanism that aligns the updates from all nodes to improve the overall stability and accuracy of the model. By focusing on the efficient distribution and synchronization of prompts, this study further refines the practical application of prompt tuning in federated environments.

Lai et al.~\cite{lai2022fedscale} offer a comprehensive benchmarking study that evaluates the performance of various federated learning models, including LLMs, across different scales and settings. The paper identifies key bottlenecks in scaling federated learning systems and proposes solutions to overcome them. Among the highlighted solutions are strategies for efficient data sampling and distribution, model compression techniques, and robust aggregation methods that can handle large-scale deployments. This benchmarking is crucial for understanding how different models and systems perform in real-world scenarios, providing a foundation for further optimization of federated learning frameworks.

Xu et al.~\cite{xu2023federated} introduce a forward gradient technique that optimizes the gradient computation and transmission in federated learning of LLMs. By calculating and sharing forward gradients instead of the full gradients, this method significantly reduces the amount of data that needs to be transmitted between nodes and the central server. This is particularly beneficial in scenarios with limited bandwidth or where data privacy is a concern. The technique also includes mechanisms for aggregating these gradients efficiently, ensuring that the model converges quickly without compromising on performance.

Together, these papers present a comprehensive view of current innovations in the efficient fine-tuning of large language models within federated learning frameworks. Starting from foundational privacy-preserving techniques, the discussion progresses through enhancements in parameter efficiency via prompt tuning, system-wide optimizations for handling scale, and innovative gradient management for improved communication efficiency. Each paper contributes to a layered understanding of how to optimize the training and deployment of LLMs across distributed networks, ensuring both performance and practical viability in real-world applications. This narrative arc not only highlights individual advancements but also illustrates the synergistic potential of combining these techniques to address the multifaceted challenges of federated learning.

In the realm of swarm intelligence within federated learning environments, the contributions of these papers are pivotal in demonstrating how decentralized systems can collaboratively refine and fine-tune large language models (LLMs) efficiently. \cite{ro2024efficient} introduces an architecture that incorporates differential privacy directly into the federated training process, reflecting the swarm intelligence concept of achieving collective goals while maintaining individual privacy. \cite{che2023federated} and \cite{zhang2023fedpetuning} further the narrative by focusing on prompt tuning—modifying a small subset of model parameters—thus reducing the complexity and computational load of federating large-scale models. This approach allows each node, akin to an agent in a swarm, to contribute more effectively and efficiently to the collective intelligence, optimizing the system’s overall learning capability.

\cite{lai2022fedscale} provides benchmarking insights that help understand the performance and system constraints when scaling up federated learning—akin to assessing the operational capacity of a swarm over varied environments. Lastly, \cite{xu2023federated} enhances the communication efficiency between nodes using a forward gradient method, significantly reducing the bandwidth needed for model updates. This mirrors a swarm’s efficiency in using limited resources to maintain robust communication across the group. Collectively, these studies exemplify how federated learning can harness swarm intelligence principles to achieve decentralized problem-solving and learning, effectively managing resources and maintaining synchronization across distributed agents to optimize collective outcomes in the training and deployment of LLMs.

\subsection{Pre-Training of LLM in Federated Learning}

\begin{table*}
\centering
\caption{Pre-Training of LLM in Federated Learning}
\vskip 6mm
\label{tab:pretraining_llm}
\begin{tabular}{c|p{4cm}|p{2cm}|p{2cm}|p{2cm}|p{4cm}}
\toprule
\textbf{No.} & \textbf{Contributions} & \textbf{Technique} & \textbf{LM used} & \textbf{Dataset} & \textbf{Limitations} \\ 
\midrule
1 & Adapts BERT pre-training to federated learning, maintaining data privacy across distributed datasets. & Federated learning adaptations, privacy enhancements & BERT & GLUE & Uneven data distribution, increased complexity in training. \\ 
\midrule
2 & Explores practical deployment of pretrained models in federated learning, addressing data heterogeneity. & Dynamic update rates, selective parameter updating & DistiBERT, BART & SST2, OntoNotes, etc & Sync issues, managing diverse data distributions. \\ 
\midrule
3 & Enhances multilingual understanding in federated learning using pretrained models across diverse languages. & Multilingual model adaptation & Multilingual BERT & MTNT & Requires careful handling of linguistic diversity, potential bias in language representation. \\ 
\midrule
4 & Reduces communication overhead by employing parameter-efficient fine-tuning techniques in federated settings. & Layer-wise relevance propagation, parameter-efficient fine-tuning & Bert-based & GLUE & Possible reduction in model accuracy, limited updates to model parameters. \\ 
\bottomrule
\end{tabular}
\end{table*}

Tian et al.~\cite{tian2022fedbert} explore the feasibility of applying federated learning to the pre-training phase of BERT (Bidirectional Encoder Representations from Transformers), a popular language model. Named FedBERT, this approach adapts the pre-training process to work across distributed data sets without compromising data privacy. The study demonstrates how federated learning can be effectively utilized to pre-train BERT with data from multiple decentralized sources, ensuring that the private data remains local while still contributing to the training of a robust model. The paper highlights modifications to the BERT training algorithm to accommodate the federated setting, including adjustments to handle the uneven distribution of data across nodes.

Focusing on the practical aspects, Agarwal et al.~\cite{agarwal2023practical} provide a detailed examination of deploying pretrained language models in a federated learning framework. It discusses the challenges and solutions related to synchronization, data heterogeneity, and maintaining model performance when the pretrained model is adapted to new datasets across distributed environments. The paper proposes several optimization techniques to enhance the efficiency and effectiveness of federated learning for pretrained models, such as dynamic update rates and selective parameter updating to cope with the diverse data distributions typically encountered in federated settings.

Addressing the challenge of multilingual contexts, Weller et al.~\cite{weller2022pretrained} introduce techniques for leveraging pretrained language models to enhance language understanding across different languages in a federated learning scenario. It explores the application of multilingual BERT models, adapting them to federated settings to improve model training across linguistically diverse data. This approach not only broadens the applicability of federated learning but also enhances the inclusivity of language technologies, allowing for effective model training without centralizing data from various language communities.

Malaviya et al.~\cite{malaviya2023reducing} address one of the significant challenges in federated learning: the high communication overhead. It presents strategies for reducing bandwidth consumption during the pre-training of language models in federated setups. By employing parameter-efficient techniques such as layer-wise relevance propagation and other fine-tuning methods that minimize the number of parameters that need to be updated and transmitted, the paper successfully decreases the network load, which is crucial for practical implementations of federated learning where bandwidth may be limited.

These papers collectively cover a spectrum of methodologies for integrating pre-training phases of large language models with federated learning principles. Starting from adapting existing models like BERT to federated settings, moving through practical deployment considerations, addressing multilingual training needs, reducing communication overhead, and finally discussing comprehensive lifecycle approaches, these studies showcase a progressive enhancement in the techniques and applications of federated learning. Each paper builds on the notion that federated learning can be effectively scaled and adapted to pre-train language models in a way that respects privacy, manages resources efficiently, and embraces linguistic diversity, thus advancing the field towards more inclusive and technologically feasible solutions.

The four papers significantly advance swarm intelligence principles in federated learning by demonstrating how large language models can be collaboratively pre-trained across distributed nodes. Each paper introduces a different facet of decentralized intelligence: \cite{tian2022fedbert} adapts BERT for privacy-preserving collaborative training, \cite{agarwal2023practical} tackles practical deployment challenges such as dynamic updating to handle data heterogeneity, \cite{weller2022pretrained} extends federated learning to multilingual contexts enhancing linguistic inclusivity, and \cite{malaviya2023reducing} reduces communication overhead with parameter-efficient techniques. Together, these studies exemplify how federated learning can emulate swarm behavior, optimizing collaborative tasks through decentralized interactions and contributing to the development of robust, scalable models without centralized control.

\begin{figure*}
\label{fig:scalable}
  \centering
  \includegraphics[width=0.80\textwidth]{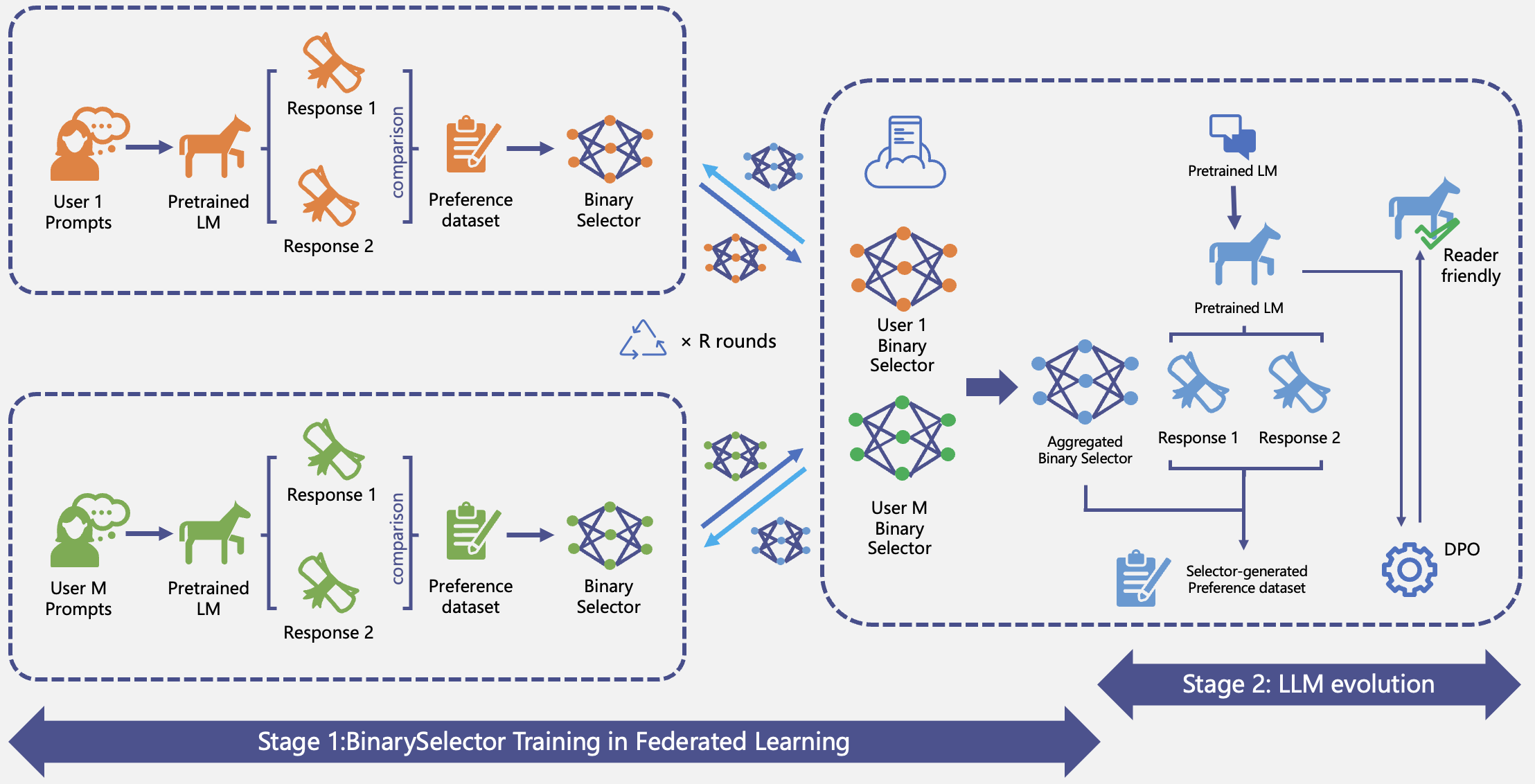}
  \caption{Scalable Federated LLMs for Swarm Intelligence }
\end{figure*}

\subsection{Scalable LLM via Federated Learning}

\begin{table*}
\centering
\caption{Scalable LLM via Federated Learning}
\vskip 6mm
\label{tab:scalable_llm}
\begin{tabular}{c|p{4cm}|p{2cm}|p{2cm}|p{2cm}|p{4cm}}
\toprule
\textbf{No.} & \textbf{Contributions} & \textbf{Technique} & \textbf{LM used} & \textbf{Dataset} & \textbf{Limitations} \\ 
\midrule
1 & Develops FATE-LLM, a framework for scalable federated learning of LLMs, with robustness against node dropout and data discrepancies. & Hierarchical aggregation & LLaMa, GPT-2, etc. & AdvertiseGen & May struggle with extremely large datasets or very high node dropout rates. \\ 
\midrule
2 & Enhances federated learning from pre-trained models using contrastive learning to maintain model quality with non-IID data. & Contrastive learning & Not specified & DomainNet, Office-10, etc. & Requires careful tuning to avoid negative impacts on convergence. \\ 
\midrule
3 & Investigates the impact of large cohorts on federated learning efficiency, proposing dynamic cohort size adjustments. & Dynamic cohort sizing & Not specified & Stack Overflow, SHAKESPEARE, etc. & Communication overhead can still be significant in very large networks. \\ 
\midrule
4 & Presents an architecture for efficiently fine-tuning LLMs in federated settings, reducing resource demands. & Selective parameter updating & DistilBERT & IMDB, AG News, etc. & Limited by the granularity of parameter selection and update mechanisms. \\ 
\midrule
5 & Discusses methodologies for training larger models in cross-device federated learning through model splitting and layered updates. & Model splitting, layered updates & LSTM, Transformer, etc. & Stack
Overflow & May face challenges in ensuring model consistency and managing update synchronization. \\ 
\bottomrule
\end{tabular}
\end{table*}

Fan et al.~\cite{fan2023fate} introduce FATE-LLM, a framework designed for training large language models using federated learning across highly distributed and heterogeneous environments. It focuses on overcoming the scalability challenges inherent in coordinating and aggregating updates across numerous devices. FATE-LLM employs a hierarchical aggregation strategy that reduces the communication burden and latency, enabling effective scalability without compromising the learning efficiency or model accuracy. The framework also includes robustness against node dropout and data discrepancies, making it particularly suitable for real-world applications where network conditions and data distributions can be highly variable.

Tan et al.~\cite{tan2022federated} explore the use of contrastive learning techniques to enhance the effectiveness of federated learning when starting from pre-trained language models. By leveraging contrastive learning, the approach aims to maximize the relevance of local updates to the global model, thus enhancing the overall model performance even when individual clients have sparse or non-IID (non-independent and identically distributed) data. The paper discusses how this technique can be particularly useful in maintaining model quality while scaling up the number of participants in a federated learning setup, addressing common issues such as catastrophic forgetting and model divergence.

Charles et al.~\cite{charles2021large} investigate the effects of increasing the cohort size (the number of participating nodes in each training round) on the efficiency and effectiveness of federated learning systems. It proposes a method for dynamically adjusting the cohort size based on real-time assessments of network bandwidth and participant availability, optimizing resource allocation and training speed. The paper provides empirical evidence that larger cohorts can lead to faster convergence and improved model performance, provided that the communication and aggregation protocols are efficiently managed.

Hilmkil et al.~\cite{hilmkil2021scaling} address the challenge of fine-tuning large pre-trained language models in a federated setting. It presents a novel architecture that allows for efficient distribution of model parameters and selective updating of those parameters most relevant to specific tasks or data types. The approach helps mitigate the high resource demands typically associated with large model fine-tuning, making it feasible to scale up federated learning to handle large language models across extensive networks of distributed nodes.

Focusing on cross-device scenarios, Ro et al.~\cite{ro2022scaling} explore methodologies to scale up the size of language models that can be effectively trained in a federated learning framework involving a wide array of devices. It introduces techniques such as model splitting and layered updates to manage the computational and memory constraints of devices, allowing even those with limited capabilities to participate in the training process. The paper emphasizes the scalability and flexibility of federated learning approaches in accommodating large models without requiring high-end hardware on each client device.

These papers collectively advance the field of federated learning by addressing various aspects of scalability when training large language models. From hierarchical aggregation and contrastive learning approaches that enhance communication efficiency and model relevance, to dynamic cohort management and architectural innovations for fine-tuning, each paper contributes to overcoming the significant challenges of scaling LLMs in distributed environments. By implementing these methodologies, federated learning can harness the power of swarm intelligence, where many distributed agents (devices) work together to solve complex problems, leading to the development of more robust and scalable language models that can operate effectively across diverse and widespread data sources.

These papers provide significant insights into how swarm intelligence principles can be applied to scale large language models (LLMs) in federated learning environments. Each paper introduces innovative techniques that enhance collaborative learning across distributed networks, reflecting key aspects of swarm behavior such as decentralized decision-making and collective problem-solving. The first paper~\cite{fan2023fate}, with its hierarchical aggregation framework, exemplifies how individual nodes can efficiently contribute to a global model without centralized oversight, much like how insects in a swarm interact locally with simple rules that lead to complex group behavior. \cite{tan2022federated}'s use of contrastive learning further leverages local computations to maintain the relevance and accuracy of the global model, ensuring that each node's update significantly contributes to the collective knowledge. \cite{charles2021large}'s dynamic cohort sizing adapts to changing network conditions and resources, optimizing the learning process similar to how swarms dynamically adjust to environmental changes. Lastly, the paper~\cite{hilmkil2021scaling} on model splitting and layered updates allows even resource-constrained devices to participate in the learning process, promoting inclusivity and resilience, akin to a swarm's ability to adapt and thrive despite individual limitations. Collectively, these studies advance the implementation of swarm intelligence in federated settings, demonstrating how decentralized, collaborative efforts can lead to robust, scalable, and efficient outcomes in the training of sophisticated models.

\begin{figure*}
\label{fig:security}
  \centering
  \includegraphics[width=0.80\textwidth]{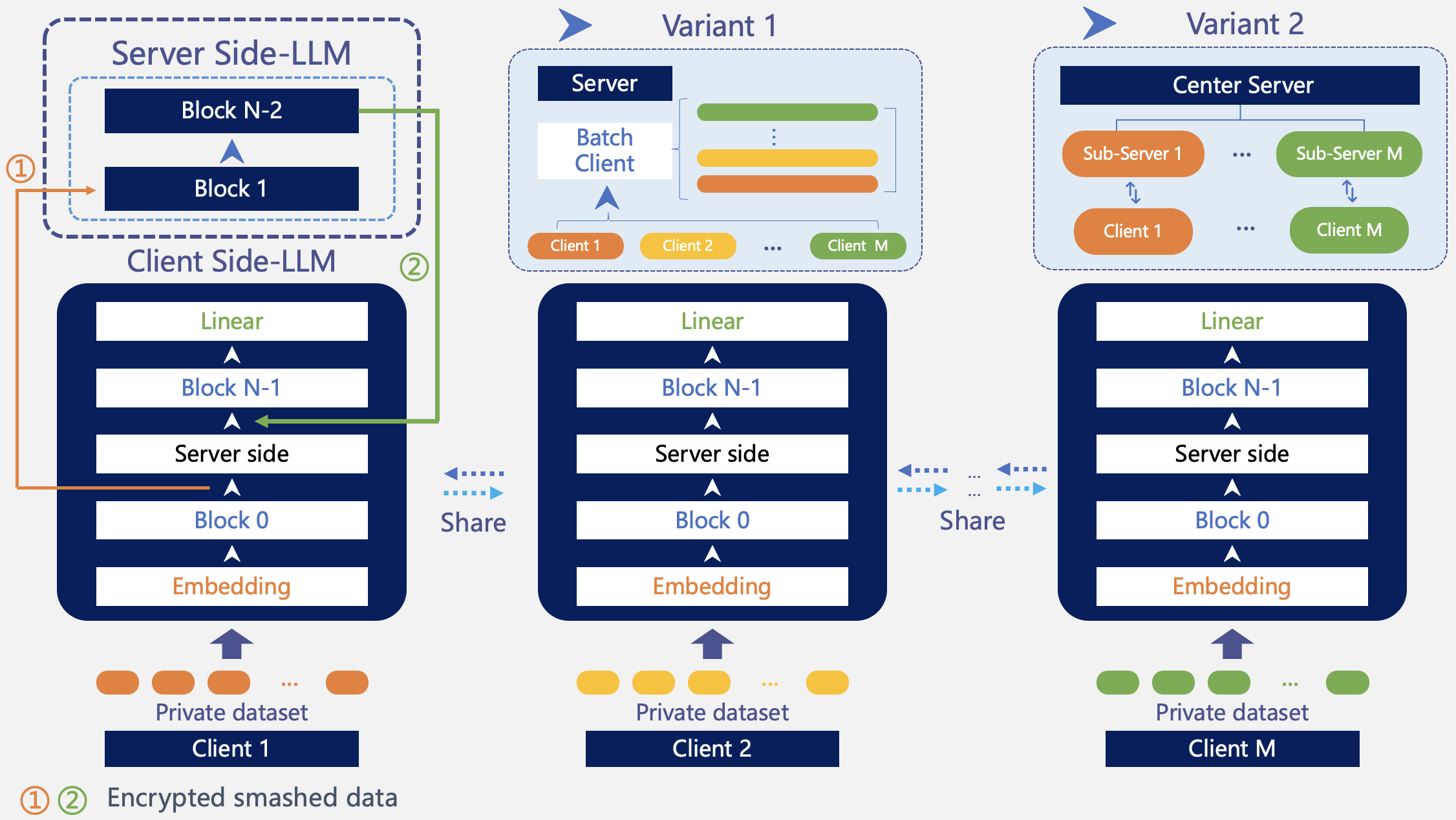}
  \caption{Secure and Private Federated LLMs for Swarm Intelligence }
\end{figure*}

\subsection{Security and Privacy in Federated LLM}

In this section, we explore the challenges and solutions associated with maintaining privacy and security while training large language models using federated learning. 

Vu et al.~\cite{vu2024analysis} provide an in-depth analysis of how privacy can be compromised during the federated training of large language models. It identifies specific vulnerabilities where sensitive data could potentially be extracted by adversaries through model inversion attacks and other inference techniques. The study systematically examines the types of information that can be leaked and under what conditions, offering insights into the limitations of current privacy-preserving methods like differential privacy. The paper suggests enhanced cryptographic measures and more robust aggregation algorithms to mitigate these risks, setting a foundational context for the need for advanced privacy-preserving techniques in federated learning.

Ye et al.~\cite{ye2024openfedllm} introduce an open-source framework specifically designed to support the development and deployment of federated learning models with an emphasis on privacy and security. This paper describes the architecture of the framework, which includes built-in support for secure multi-party computation (SMPC) and homomorphic encryption, technologies that allow computations to be performed on encrypted data. The framework aims to provide a practical solution to the privacy concerns highlighted in the first paper, facilitating the secure aggregation of updates from multiple clients without exposing their raw data, thereby enhancing the overall security of the federated learning process.

Building on the concerns about privacy leaks, Gupta et al.~\cite{gupta2022recovering} demonstrate a practical attack scenario where sensitive information is extracted from a federated language model. The researchers show how certain training techniques and model configurations can inadvertently lead to the memorization of private text, which can then be recovered by malicious participants. The paper tests various scenarios and configurations, providing empirical evidence of the risks involved. It also discusses potential countermeasures, such as more stringent data sanitization and the use of privacy-preserving architectures, to prevent such vulnerabilities.

Thakkar et al.~\cite{thakkar2021understanding} complement the third by further investigating how language models, especially those trained under federated learning conditions, may unintentionally memorize and expose sensitive data. It delves into the mechanics of memorization in neural networks, exploring how different factors like model size, dataset diversity, and training duration affect the risk of data leakage. The study proposes a set of best practices and modifications to the training algorithms that can help reduce the likelihood of such memorization without significantly impacting model performance.

Together, these papers create a layered understanding of the privacy and security challenges in federated learning of LLMs. Starting from identifying potential privacy leaks, moving to a framework designed to address these leaks through encryption and secure computation, and finally, demonstrating practical attack vectors and offering solutions to mitigate these issues, the narrative crafted by these studies underscores the complexity of securely training language models in a distributed manner. Each paper builds on the insights of its predecessors, collectively advancing the field towards more secure and private federated learning environments. This progression not only highlights the vulnerabilities but also charts a path toward resolving them, ensuring that federated learning can be safely employed in sensitive applications.

The papers in this section contribute to the principles of swarm intelligence in the context of federated learning for large language models by addressing complex challenges related to security and privacy—key aspects for collaborative and distributed systems. First, by analyzing potential privacy leaks, the initial paper lays the groundwork for understanding vulnerabilities akin to assessing environmental risks in a swarm. \cite{ye2024openfedllm} introduces a robust framework with advanced encryption methods, enhancing collective security without centralized control, much like a swarm's distributed processing enhances collective resilience. \cite{gupta2022recovering} showcases practical attack simulations and mitigations, reinforcing the swarm's adaptability and response to threats. Finally, \cite{thakkar2021understanding}'s investigation into unintended data memorization offers strategies to minimize such risks, promoting safer learning environments. Collectively, these studies embody swarm intelligence by improving the collective learning and decision-making processes in a decentralized, secure, and private manner, ensuring the swarm (network of nodes) remains robust against internal and external adversarial influences.

\begin{table*}
\centering
\caption{Security and Privacy in Federated LLM}
\vskip 6mm
\label{tab:security_privacy_fllm}
\begin{tabular}{c|p{4cm}|p{2cm}|p{2cm}|p{2cm}|p{4cm}}
\toprule
\textbf{No.} & \textbf{Contributions} & \textbf{Technique} & \textbf{LM used} & \textbf{Dataset} & \textbf{Limitations} \\ 
\midrule
1 & Analyzes potential privacy leaks and suggests enhanced cryptographic measures and robust aggregation algorithms. & Cryptographic measures, robust aggregation & GPT-2, RoberTa, etc. & IMDB, Yelp, etc. & May involve high computational overhead and complexity in implementation. \\ 
\midrule
2 & Introduces an open-source framework incorporating SMPC and homomorphic encryption to enhance security. & SMPC, homomorphic encryption & GPT-4, LLaMa-2, etc. & Alpaca, UltraFeedback, etc.  & Potential performance degradation due to encryption overhead. \\ 
\midrule
3 & Demonstrates practical attacks for recovering private text and discusses countermeasures. & Attack simulation, countermeasure development & GPT-2 & WikiText-103 & Countermeasures may not fully prevent leakage and could affect model efficiency. \\ 
\midrule
4 & Investigates unintended memorization and proposes best practices to reduce data leakage. & Analysis of memorization, training modifications & Bert-based & Stack Overflow & Best practices may not be universally effective across all model types and scenarios. \\ 
\bottomrule
\end{tabular}
\end{table*}

\section{Open Challenges and Future Directions}

In this section, we show the learned lessons, open challenges, and future research directions derived from the review and analysis.

\subsection{Open Challenges and Future Directions in LLM Architectures in Federated Settings}

\textbf{How can we adapt LLM architectures to accommodate diverse computational capacities across federated nodes?} Current LLMs are generally designed for centralized, high-performance computing environments, making their deployment in federated settings challenging. One approach is to design adaptive architectures that scale model complexity based on each node’s available resources. For example:

     Layer-skipping techniques could allow smaller or less capable devices to bypass certain layers in deep models, focusing on critical layers for inference, while more capable nodes handle the full architecture.
     
     Modular model structures where nodes with limited resources only compute essential components (e.g., embedding layers) while offloading deeper layers to edge or cloud infrastructure. Another approach could involve using lightweight proxy models on resource-constrained nodes to participate in training with minimal computation.

\textbf{How can we achieve efficient model aggregation without incurring high communication costs in large-scale federated networks?} As federated networks scale, the communication costs of aggregating model updates from numerous nodes can be prohibitively high. Developing hierarchical aggregation mechanisms can alleviate this by organizing nodes into tiers, where local aggregations occur before updates are sent to a central server.

     Edge-based aggregation tiers could further reduce communication by using intermediary nodes for partial aggregation, reducing the frequency and volume of data sent to the central server.
     
     Quantized and compressed model updates are another promising area. Here, nodes can send reduced precision or partially pruned updates, thus reducing data transfer needs without a significant loss in model accuracy.

\textbf{How can federated LLMs effectively manage model and data divergence due to non-iid data distributions across nodes?} Non-iid data distributions across federated nodes create challenges in training, often leading to biased models. Addressing this requires innovative aggregation and model adjustment techniques:

     Adaptive weighting of updates: Develop methods that assign weights to node updates based on similarity to the global data distribution. This could involve clustering nodes based on data profiles, so model updates from similar distributions are aggregated separately before global merging.
     
     Divergence-aware training algorithms: New algorithms could incorporate divergence checks, allowing the model to adapt parameters dynamically based on the node’s data distribution. Exploring approaches like federated multi-task learning where each node’s model partially diverges based on data uniqueness could also help maintain robustness.

\textbf{How can we design energy-efficient LLM architectures for resource-constrained federated environments?} To mitigate the high energy demands of LLMs, future models could focus on parameter-efficient techniques that reduce the computational load:

     Sparse and low-rank modeling: Sparse representations or low-rank approximations reduce the number of active model parameters without sacrificing performance. This can be done by pruning layers or enforcing sparsity during training.
     
     Green federated architectures: New architectures could integrate energy monitoring and adaptive energy-use policies that dynamically adjust computation based on each node’s available energy resources, limiting participation in training rounds if necessary to save energy.
     
     Model distillation: Using smaller, distilled versions of large models that retain core functionalities for specific tasks can significantly reduce energy use. Research into task-specific distillation for federated LLMs is an effective way to make LLM deployment feasible on lower-energy devices.

\textbf{How can federated LLM architectures improve interoperability between edge and cloud infrastructures to leverage their complementary strengths?} Designing hybrid architectures that operate seamlessly across edge and cloud infrastructures could enhance the functionality and scalability of federated LLMs.

     Hybrid federated models: One approach is to create hybrid models that allow computationally intensive processes to run on the cloud, while edge devices focus on localized or preliminary tasks (e.g., feature extraction). Nodes could switch dynamically between edge and cloud depending on task complexity and resource availability.
     
     Resource-aware task allocation: Develop systems where tasks are automatically assigned based on resource availability, sending large-scale computations to the cloud when edge resources are limited.
     
     Cross-infrastructure synchronization: Federated LLMs could benefit from mechanisms that allow seamless model synchronization between edge and cloud, ensuring consistency and minimizing delays in model updates or inference responses.

\subsection{Open Challenges and Future Directions in Efficient Fine-Tuning of Federated LLM}

\textbf{How can we develop fine-tuning techniques that balance privacy, efficiency, and model performance in federated LLMs?} Privacy-preserving mechanisms such as differential privacy and secure multi-party computation can limit model performance due to added noise or increased complexity. Future research could explore:
     
     Lightweight differential privacy methods: Instead of adding uniform noise, adaptive noise addition techniques that consider the model’s sensitivity to data can help minimize privacy-related performance losses.
     
     Federated privacy modules: Developing modular privacy-preserving components within federated architectures allows nodes to use privacy techniques based on their specific privacy requirements and computational abilities.
     
     Hybrid privacy approaches: Combining differential privacy with other privacy-preserving techniques, such as homomorphic encryption, could create a layered approach that tailors privacy levels based on node-specific data sensitivities.

\textbf{What are the most effective methods to achieve parameter efficiency in federated fine-tuning without compromising model accuracy?} Large models can be challenging to fine-tune due to resource constraints on federated nodes. Future directions could include:

     Prompt tuning and adapter modules: These methods adjust only a small subset of model parameters rather than the entire model. Nodes would use small, localized adapters to improve efficiency, with the main model remaining fixed.
     
     Sparse and selective fine-tuning: Research could focus on selectively fine-tuning only the most relevant layers or parameters of a model, dynamically identifying and targeting those areas based on the specific data characteristics of each node.
     
     Federated parameter sharing: A strategy in which nodes share only essential parameter updates, minimizing the transmission of non-impactful parameters, could further reduce resource requirements while preserving accuracy.

\textbf{How can federated LLMs manage non-iid data efficiently during the fine-tuning process?} Non-iid data across federated nodes complicates fine-tuning as it may lead to overfitting or poor generalization in the global model. Approaches could include:

     Localized fine-tuning with periodic global synchronization: Nodes could perform local fine-tuning specific to their data, with occasional synchronization to align with the global model.
     
     Clustered fine-tuning groups: Grouping nodes with similar data distributions for clustered fine-tuning can improve convergence by enabling models that are optimized for similar tasks or data profiles.
     
     Dynamic weighting of updates: Introducing adaptive weighting mechanisms that prioritize updates from nodes with data distributions more similar to the target application could enhance performance for non-iid scenarios.

\textbf{How can we optimize communication overhead in federated fine-tuning for large-scale LLMs?} Communication costs can become prohibitive, especially for large models in federated settings. Potential strategies include:

     Forward gradient updates: Reducing the data sent by transmitting only forward gradients, which simplifies updates without requiring complete backpropagation at each node.
     
     Compression and sparsification of updates: Techniques like gradient sparsification or quantization reduce the volume of data transmitted while retaining the core information needed for model updates.

     Asynchronous fine-tuning protocols: Implementing asynchronous update protocols where nodes transmit updates based on local conditions and bandwidth availability can help balance communication loads across the network.

\textbf{What techniques can improve adaptive learning across diverse federated environments to support real-time fine-tuning?} Real-time adaptive learning could help federated LLMs stay relevant to continuously changing data distributions. Future approaches could include:

     Adaptive learning rate optimization: Techniques that dynamically adjust learning rates based on node-specific convergence rates and data characteristics can enhance learning stability and efficiency.
     
     Federated learning decay methods: Introducing learning decay for nodes that receive frequent updates, helping them prioritize updates from nodes with less frequent yet impactful data, could improve adaptive learning across dynamic federated environments.
     
     Edge-based update caching: Nodes could store recent updates and synchronize them only when significant model changes are detected, improving efficiency and allowing real-time adaptation without overwhelming network resources.

\subsection{Open Challenges and Future Directions in Pre-Training of LLM in Federated Learning}

\textbf{How can we effectively pre-train LLMs across federated nodes with limited access to high-quality and diverse data?} Federated settings often involve data that is both non-iid and diverse in quality. Strategies to address this include:

     Synthetic data augmentation: Developing methods to generate synthetic data that reflect node-specific characteristics could help compensate for the lack of diverse and high-quality data.
     
     Collaborative data enrichment: Nodes with limited data could be paired with nodes that have high-quality, relevant datasets, allowing for controlled data exchange or guided knowledge transfer to enhance training.
     
     Federated curriculum learning: Implement a curriculum that gradually introduces complex data during pre-training, aligning nodes based on data complexity and quality levels to improve model robustness.

\textbf{What methods can be developed to handle high communication costs during federated pre-training?} Pre-training large LLMs in federated environments requires significant data exchange. Future research could explore:

     Layer-wise or selective parameter updates: Rather than sending updates for the entire model, nodes could transmit updates for only the most relevant layers or parameters during the pre-training phase.
     
     Federated model compression: Investigate advanced compression techniques that reduce the communication footprint of model updates, such as quantization or pruning applied to updates before transmission.
     
     Hierarchical update aggregation: Group nodes in clusters to perform intermediate aggregations locally before sending the consolidated updates to the global server, thereby reducing the overall communication load.

\textbf{How can federated pre-training account for language or domain-specific variations across nodes?} In multilingual or domain-specific federated environments, pre-training faces the challenge of learning from highly varied data. Solutions may include:

     Multi-branch model architectures: Develop multi-branch models where each branch specializes in a particular language or domain, with shared parameters for common knowledge and separate branches for specialization.
     
     Localized model tuning: Incorporate localized fine-tuning layers within the pre-trained model, allowing nodes to emphasize domain-specific knowledge while contributing to the shared global model.
     
     Cross-lingual or domain-specific embeddings: Nodes can maintain domain-specific or language-specific embeddings, which could then be aligned with the global embedding space periodically, promoting coherence without sacrificing specialization.

\textbf{How can pre-training models in federated learning handle data privacy and regulatory constraints?} Privacy concerns are heightened in the pre-training stage, where large volumes of data are often required. Future research could focus on:

     Federated differential privacy techniques: Develop adaptive differential privacy techniques that adjust noise levels based on the sensitivity of the data and the stage of training, balancing privacy with model performance.

     Secure multi-party computation (SMPC) for pre-training: Incorporate SMPC methods to allow federated nodes to contribute pre-training data securely without direct data exposure.
     
     Federated zero-knowledge proofs: Explore the use of zero-knowledge proofs to validate pre-training computations while preserving the privacy of the data, ensuring compliance with stringent regulatory requirements.

\textbf{How can we manage model stability and convergence across nodes during federated pre-training?} Convergence in federated pre-training is challenging, particularly as nodes may have differing data distributions. Potential approaches include:

     Federated model checkpointing: Implement periodic checkpointing that captures intermediate model states, allowing nodes to revert to stable checkpoints if divergence is detected.
     
     Adaptive synchronization frequencies: Adjust synchronization intervals dynamically based on the convergence rate of each node, with nodes that struggle to converge updating more frequently.
     
     Federated ensemble methods: Employ ensemble learning techniques to stabilize training by aggregating models from different nodes into a federated ensemble, allowing the global model to better generalize across diverse data distributions.

\subsection{Open Challenges and Future Directions in Scalable LLM via Federated Learning}

\textbf{How can we scale federated LLMs to handle an increasing number of nodes without overwhelming network and computational resources?} Scaling federated LLMs requires innovations that minimize network load and computational strain. Promising approaches include:
    
     - Hierarchical and layered aggregation: Organize nodes into hierarchies, where initial aggregations occur within smaller groups before being shared with a central server. This reduces the amount of data transmitted and allows for scalable participation across large networks.
    
     - Edge-based preprocessing and filtering: Incorporate preprocessing at the edge, allowing nodes to filter or pre-aggregate updates before sending them to the server. This approach can help reduce data redundancy and improve communication efficiency.
    
     - Adaptive participation protocols: Develop adaptive protocols that limit the participation of nodes based on network conditions and model convergence needs, thereby reducing resource demand during peak times.

\textbf{How can we address model consistency and synchronization issues as federated LLMs scale?} As the number of nodes grows, maintaining consistency and synchronizing updates becomes challenging. Solutions could include:
    
     Asynchronous model updates: Design protocols that allow nodes to asynchronously update the model, with a central reconciliation process to ensure that updates remain consistent and do not diverge excessively.
    
     Dynamic weighting of node contributions: Assign higher weights to updates from nodes with higher-quality data or better alignment with the global model, helping to stabilize the model as more nodes participate.
   
     Federated caching mechanisms: Implement caching on edge nodes or intermediate servers to temporarily store updates. This can help nodes sync efficiently and mitigate the delays in model aggregation.

\textbf{What methods can make federated LLMs more computationally feasible for resource-limited nodes in large networks?} Many federated nodes may be resource-constrained, making participation in large-scale LLM training challenging. Strategies to address this include:
    
     Model distillation and pruning: Implement techniques like model distillation, where a smaller, distilled version of the LLM is shared with resource-constrained nodes, allowing them to contribute to training with fewer computational requirements.
    
     Selective layer updating: Enable nodes to update only a subset of layers (such as the final few layers or embeddings) rather than the entire model, reducing computational load while still contributing effectively to learning.
    
     Federated model partitioning: Distribute different parts of the model across nodes, so each node focuses on specific layers or sections. This method allows large-scale models to be trained without requiring each node to handle the full model.

\textbf{How can federated learning frameworks ensure robust performance when scaling LLMs across nodes with non-iid data distributions?} Non-iid data can lead to instability and inconsistencies in federated learning as models scale. Future work can focus on:
    
     Federated clustering: Group nodes based on data similarities so that each cluster can train a model variant optimized for its specific data distribution, with occasional merging to create a generalized global model.
    
     Adaptive local training epochs: Allow nodes to train for different numbers of epochs based on their data variance, ensuring nodes with high variability train longer before syncing with the global model.
   
     Domain-specific layers: Develop modular architectures where specific layers are tuned for different data distributions, allowing nodes to fine-tune only domain-specific layers and share core model parameters globally.

\textbf{How can scalability be achieved while ensuring data privacy and security across highly distributed federated LLM networks?} Privacy becomes increasingly critical as federated networks scale. Potential research directions include:
    
     Federated encryption and differential privacy: Apply encryption and differential privacy selectively, focusing on layers or parameters with the highest risk of data leakage to minimize resource costs while maintaining security.
    
     Secure multi-party computation (SMPC) with hierarchical structure: Employ SMPC protocols in a hierarchical setting, where data can be aggregated within smaller groups securely before sharing more broadly, reducing computational overhead.
  
     Decentralized privacy management: Design privacy-preserving protocols that allow individual nodes to handle sensitive data autonomously, with encrypted summaries shared to a central server to ensure privacy without exposing raw data.

\subsection{Open Challenges and Future Directions in Security and Privacy in Federated LLM}

\textbf{How can federated LLMs prevent privacy leakage while allowing effective model training?} Protecting privacy while enabling effective training requires advanced privacy-preserving mechanisms:

     Advanced differential privacy techniques: Implement differential privacy with adaptive noise levels, where sensitive data receives higher protection while general information has lower noise, helping maintain performance.
     
     Federated secure multi-party computation (SMPC): Enhance SMPC protocols specifically for LLMs, ensuring that all computations occur over encrypted data without compromising model accuracy.
     
     Federated data masking: Use data masking or perturbation for high-risk data fields, reducing the chance of privacy leakage in LLMs trained across diverse and sensitive datasets.

\textbf{What methods can be used to detect and defend against adversarial attacks on federated LLMs?} Adversarial attacks can compromise federated LLMs, necessitating robust detection and defense methods:

     Anomaly detection mechanisms: Develop real-time anomaly detection systems that can identify and flag suspicious updates or node behaviors, leveraging statistical and machine learning methods to detect abnormal patterns in model updates.
     
     Federated adversarial training: Incorporate adversarial training across federated nodes, with nodes introducing synthetic adversarial examples to make the model more resilient to attacks.
     
     Dynamic threat assessment and response: Design dynamic threat detection protocols that assess the risk level of each node based on its history and data sensitivity, allowing the system to adjust privacy protections accordingly.

\textbf{How can federated learning architectures minimize unintended memorization to protect sensitive data?} LLMs can inadvertently memorize sensitive data, posing privacy risks. Techniques to mitigate this include:

     Regularization techniques for federated setups: Implement dropout, weight decay, or other regularization methods that reduce memorization, encouraging the model to generalize rather than memorize specific data points.
     
     Selective forgetting protocols: Introduce machine unlearning techniques in federated LLMs, allowing nodes to identify and “forget” specific data points that may pose privacy risks.
     
     Federated differential memorization control: Explore frameworks that analyze model parameters for sensitive information retention, selectively adjusting training steps to reduce memorization risks.

\textbf{How can federated LLMs ensure secure data exchange and aggregation across distributed nodes?} Securely aggregating data from distributed nodes requires robust encryption and secure data handling:

     Homomorphic encryption for secure aggregation: Implement homomorphic encryption to allow nodes to perform encrypted aggregations, reducing risks of data exposure during transfer.
     
     Zero-knowledge proofs (ZKPs) for verification: Develop ZKPs in federated learning, enabling nodes to prove they’ve performed computations correctly without revealing underlying data, enhancing trust across nodes.
     
     Encrypted aggregation frameworks: Design frameworks where nodes encrypt their data contributions, with partial aggregations at intermediary nodes that maintain full encryption, preserving data security throughout the aggregation process.

\textbf{What frameworks can enforce data usage policies and regulatory compliance in federated LLMs?} As federated LLMs often involve sensitive data, ensuring compliance with regulations (e.g., GDPR) is crucial:

     Federated audit trails: Create an audit system that tracks data use and model updates at each node, allowing for transparency and compliance checks with minimal overhead.
     
     Privacy-preserving logging mechanisms: Implement logging that captures node interactions and updates in a secure, anonymized format to enable regulatory review without compromising privacy.
     
     Policy-aware federated learning frameworks: Develop frameworks that automatically enforce data retention, access, and deletion policies across nodes, with clear guidelines and checks for regulatory compliance.

\section{Conclusion}
This survey has examined the integration of Federated Learning (FL) and Large Language Models (LLMs), providing a comprehensive overview of the architectural foundations, performance characteristics, and security considerations of this emerging paradigm. We have shown that FL offers a promising solution for training and deploying LLMs across decentralized and privacy-sensitive environments, enabling scalable, compliant, and efficient model development without compromising data ownership.

Our exploration highlighted a variety of approaches aimed at improving model robustness, communication efficiency, and security, while also identifying open challenges such as handling data heterogeneity, ensuring consistent convergence, and supporting privacy-preserving mechanisms like machine unlearning. Despite these challenges, the synergy between FL and LLMs presents a forward-looking direction for building AI systems that are both powerful and trustworthy.

As research and deployment efforts continue to expand, the federated training of LLMs is poised to reshape the future of distributed AI, empowering applications in healthcare, finance, education, and beyond with adaptive, secure, and ethically grounded language technologies.

\vskip 6mm

\renewcommand\refname{\zihao{5}\textbf{References}}

\bibliographystyle{IEEEtran}
\bibliography{ref}

\end{document}